\documentclass{article}

\usepackage{arxiv}

\usepackage[utf8]{inputenc} 
\usepackage[T1]{fontenc}    
\usepackage{hyperref}       
\usepackage{url}            
\usepackage{booktabs}       
\usepackage{amsfonts}       
\usepackage{nicefrac}       
\usepackage{microtype}      
\usepackage{lipsum}
\usepackage{amsmath}
\usepackage{graphicx}
\graphicspath{ {./images/} }

\title{Zero Shot Composed Image Retrieval}

\author{
 Santhosh Kakarla \\
  George Mason University\\
  \texttt{skakarl3@gmu.edu} \\
   \And
 Gautama Shastry Bulusu Venkata \\
  George Mason University\\
  \texttt{sbulusuv@gmu.edu} \\
}

\begin{document}
\maketitle
\begin{abstract}
Composed image retrieval (CIR) enables users to perform fine‑grained visual edits—such as “turn the dress blue” or “remove stripes”—by combining a reference image with a textual instruction. Zero‑shot CIR methods using pretrained multimodal encoders typically encode text and image separately, yielding $\sim$20–25\% Recall@10 on benchmarks like FashionIQ. To enhance performance, we apply a BLIP2‑based fine‑tuning strategy with a Q‑Former to fuse image and text into a unified representation, boosting Recall@10 to 45.6\% (shirt), 40.1\% (dress), and 50.4\% (toptee) and average Recall@50 to 67.6\%. Building on this, we implement Retrieval‑DPO with CLIP’s separate text and vision encoders: each text prompt is encoded via the CLIP transformer, positives and FAISS‑mined hard negatives are encoded via CLIP’s image encoder, and a Direct Preference Optimization loss enforces a margin between their similarity scores. Despite extensive tuning of the DPO scaling factor, index parameters, and sampling strategies, this approach achieves only 0.02\% Recall@10—far below zero‑shot and prompt‑tuned baselines—due to (1) lack of joint image–text fusion, (2) misalignment between margin‑based loss and top‑$K$ metrics, (3) poor negative quality, and (4) frozen visual and transformer layers. These findings underscore the necessity of true multimodal fusion, ranking‑aligned objectives, and refined negative sampling for effective preference‑based CIR. 
\end{abstract}


\section{Introduction}
Composed image retrieval (CIR) enables users to issue semantically rich visual search queries by specifying a reference image alongside a textual modification—examples include “turn the dress blue,” “remove the floral pattern,” or “add stripes to the shirt.” Zero‑shot CIR (ZSCIR) methods process these queries by feeding images and text independently into pretrained multimodal encoders (e.g., CLIP) and computing cosine similarity between the two modalities. This pipeline requires no task‑specific training data, making it highly flexible; however, it struggles to interpret and combine visual and linguistic nuances. On the FashionIQ benchmark, which contains real‑world product images and human‑annotated edit instructions, ZSCIR delivers only ~20–25\% Recall@10. Such limited performance hinders CIR’s utility in critical applications—like e‑commerce personalization, where customers expect precise product variations; digital design prototyping, which demands fine‑grained control over visual edits; and content curation or moderation, where semantics-driven filtering is essential.

To overcome these limitations, we adopt a fine‑tuning strategy within the BLIP2 framework, introducing a lightweight Q‑Former module that learns to fuse visual and textual inputs into a unified representation. During fine‑tuning, the Q‑Former consumes both the embedding of the reference image and the token embeddings of the edit instruction, producing a single, dynamic prompt vector. This fused prompt is then compared against candidate images via a contrastive retrieval loss, aligning it more closely with the target image and distancing it from negatives. Without requiring paired examples of edited images, this approach adapts the generic multimodal encoder to the CIR task, improving its sensitivity to subtle attribute changes and object manipulations. Empirically, fine‑tuned BLIP2 achieves Recall@10 of 45.6\% for shirt edits, 40.1\% for dress edits, and 50.4\% for toptee edits, with an average Recall@50 of 67.6\% on FashionIQ.

Building on these gains, we investigate Retrieval‑DPO for CIR, a retrieval‑augmented preference optimization framework that seeks to inject user‑style ranking signals directly into the retrieval model. Unlike prompt‑tuning, this method operates on separate CLIP text and vision encoders: each caption is encoded via the CLIP transformer into a prompt embedding, while positive and FAISS‑mined hard negative images are encoded by the CLIP vision encoder. We retrieve the hardest negatives for each query using a FAISS index over the gallery and apply a Direct Preference Optimization loss—maximizing the log‑odds margin between positive and negative similarity scores. Despite exhaustive hyperparameter tuning (including the DPO scaling factor $\beta$, index configurations, and negative‑sampling strategies) and mixed‑precision training, Retrieval‑DPO yields only 0.02\% Recall@10. We attribute this stark underperformance to four factors: (1) the absence of true multimodal fusion limits the model’s capacity to learn cross‑modal relations; (2) DPO’s margin‑based objective does not directly optimize top‑$K$ ranking metrics; (3) FAISS‑mined negatives are either too trivial or too ambiguous, weakening the preference signal; and (4) freezing both visual and transformer layers restricts representational adaptation. These findings illuminate critical gaps in naive preference‑based approaches and underscore the need for integrated fusion architectures, ranking‑aligned loss functions, and refined negative‑sampling strategies in future multimodal preference learning research.

\section{Related Work}
\label{sec:headings}
Contrastive Language--Image Pretraining (CLIP) by Radford et al.\ (2021) introduced a dual-encoder framework that learns aligned image and text representations via a contrastive loss, enabling effective zero-shot transfer across a variety of vision-language tasks, including composed image retrieval (CIR). Building on CLIP’s paradigm, BLIP2 by Li et al.\ (2023) employs a lightweight Q-Former to learn query embeddings that fuse frozen vision encoders with large language models, significantly improving multimodal fusion for retrieval and other downstream tasks. Direct Preference Optimization (DPO) by Rafailov et al.\ (2023) reformulates reinforcement learning from human feedback into a supervised, margin-based loss that aligns model outputs with human preferences. Retrieval-DPO extends this idea to retrieval settings by mining hard negatives using FAISS and applying a joint-preference loss to the similarity scores between query and image embeddings. Collectively, these methods motivate our investigation into prompt-tuned CIR and preference-based retrieval optimization.

\section{Methodology}

Composed Image Retrieval (CIR) combines a reference image with a natural-language edit instruction to retrieve a “target” image reflecting the specified modification. Traditional CIR approaches required fully supervised training on triplets of (reference, edit, target), which is labor-intensive to collect. Zero-Shot CIR (ZSCIR) circumvents this by leveraging models like CLIP, which independently embed images and text into a shared vector space via a contrastive pretraining objective (InfoNCE, Rusak et al. 2024). Retrieval is then performed by computing cosine similarity between the text-encoded instruction and image embeddings. While ZSCIR requires no task-specific data, it struggles with fine-grained edits—achieving only 20–25\% Recall@10 on FashionIQ—because it cannot explicitly model the interaction between image content and textual modifications.

\subsection{Dataset Preparation}

The FashionIQ benchmark (Wu et al. 2019) serves as our evaluation corpus. It comprises three apparel categories—\emph{shirt} (~17 K pairs), \emph{dress} (~16 K pairs), and \emph{toptee} (~18 K pairs)—each drawn from real-world e-commerce product listings. Each reference–target pair is annotated with one to five human-written captions describing the desired visual modification (e.g., “make the dress red”, “add stripes to the shirt”, “remove floral patterns”), resulting in over 50 000 unique edit instructions across the dataset.

FashionIQ provides official train, validation, and test splits via JSON files (\texttt{split.<category>.<split>.json}), with approximately 80\% of pairs for training, 10\% for validation, and 10\% for testing. These splits ensure no overlap of reference or target images. We use the training split for model fine-tuning, the validation split for hyperparameter selection and early stopping (monitoring Recall@10), and the test split only for final performance reporting.

All images are preprocessed uniformly: loaded in RGB, resized to \(224\times224\) with bicubic interpolation, center-cropped, converted to PyTorch tensors, and normalized using CLIP’s statistics (mean = [0.481, 0.457, 0.408]; std = [0.268, 0.261, 0.276]). Text edit instructions are prefixed with \texttt{<|image|>} to signal multimodal input and tokenized to a fixed length (77 tokens for CLIP; 32 learnable queries for BLIP2’s Q-Former‐based prompt‐tuning), with truncation or padding as necessary. This consistent preprocessing pipeline ensures fair comparisons across zero-shot, prompt-tuned, and preference-optimized retrieval methods.

\subsection{Fine tuning with BLIP2’s Q‑Former}

To effectively merge visual context and textual edits for composed image retrieval, a small fusion module known as the Q‑Former is introduced between the frozen image and language components of the BLIP2 architecture. The process begins by encoding each reference image through a pretrained Vision Transformer, which outputs a grid of patch embeddings that capture localized features such as color, texture, and shape. These patch embeddings alone cannot account for the user’s desired modification, so the Q‑Former employs a fixed set of learnable query vectors that attend across the entire grid, producing a compact set of fused tokens that highlight image regions most relevant to the forthcoming edit.

Next, these fused visual tokens are projected via a trainable linear layer into the same embedding space as the language model’s tokens. Simultaneously, the textual edit instruction—prefixed by a special image marker—is tokenized and embedded by the frozen language transformer’s input layer. By concatenating the projected visual queries with the text token embeddings into a single sequence, the model creates a true multimodal input. This combined sequence is then processed by the frozen language backbone, whose self‑attention layers further integrate information across modalities. The hidden state at a designated classification position (often the first token) is extracted as the final multimodal prompt embedding, encapsulating both the reference image’s visual cues and the semantics of the requested edit.

Training this fusion head leverages a contrastive retrieval objective. In each batch, prompt embeddings are paired with their corresponding target image embeddings—also produced by the frozen vision encoder—and contrasted against all other images in the batch. The InfoNCE loss encourages each prompt to maximize cosine similarity with its true target while minimizing similarity to distractor images, using a temperature hyperparameter to modulate the hardness of negatives. Crucially, only the Q‑Former and its projection parameters are updated; the main vision and language backbones remain frozen, preserving their general-purpose capabilities and dramatically reducing the number of trainable parameters.

Optimization proceeds with the AdamW algorithm, typically using a learning rate of 1e‑4 and a modest weight decay. Mini‑batches of around 128 triplets strike a balance between stable gradient estimation and GPU memory constraints. Training generally converges in 5–10 epochs, with early stopping determined by performance on a held‑out validation set—specifically, Recall@10 on composed retrieval tasks. This targeted prompt‑tuning strategy adapts the large, frozen BLIP2 model to the demands of composed image retrieval, yielding significant improvements over zero‑shot baselines without the need for extensive paired data.

Training uses an InfoNCE contrastive loss:
\[
  \mathcal{L}_{\mathrm{InfoNCE}}
  = -\frac{1}{B} \sum_{i=1}^B
    \log \frac{\exp(\langle z_i, z_i^+\rangle / \tau)}
             {\sum_{j=1}^B \exp(\langle z_i, z_j^-\rangle / \tau)},
\]
where \(z_i\) is the \(i\)th prompt embedding, \(z_i^+\) the matching target image embedding, \(z_j^-\) the \(j\)th negative embedding, and \(\tau\) the temperature hyperparameter. Only the Q-Former and its projection layer are updated (AdamW, learning rate \(1\times10^{-4}\), weight decay \(1\times10^{-2}\)). Mini-batches of 128 triplets and mixed-precision training ensure convergence in 5–10 epochs, with early stopping based on validation Recall@10.

\subsection{Retrieval‑Augmented Direct Preference Optimization (Retrieval‑DPO)}

The Retrieval‑DPO framework begins by constructing an efficient nearest‑neighbor index over the entire image gallery. All images from the training and validation splits are preprocessed—resized to \(224\times224\), center‑cropped, converted to tensors, and normalized—and then encoded through CLIP’s frozen vision encoder. Their 1,024‑dimensional feature vectors are collected into a NumPy array, L2‑normalized, and added to a \texttt{faiss.IndexFlatIP} structure for rapid inner‑product searches. To avoid re-encoding on each run, both the FAISS index file and a pickled mapping from index rows to image filenames are cached and loaded on subsequent executions.

Training data are drawn from the FashionIQ train split, where each entry provides a caption and its matching target image filename. For each example, the caption is prefixed with \texttt{<|image|>}, tokenized via CLIP’s tokenizer, and converted to token IDs. The true target image is encoded to obtain the positive embedding, and a hard negative is mined by querying the FAISS index for the top \(k=50\) nearest neighbors and selecting the first neighbor whose index differs from the positive’s. This yields semantically or visually similar distractors for meaningful contrast.

During training, the tokenized caption is passed through CLIP’s frozen text embedding layer and transformer under mixed precision to produce a prompt embedding; positive and negative images are likewise encoded. All embeddings are L2‑normalized, and similarity scores \(s_i^+\) and \(s_i^-\) are computed via dot product and scaled by CLIP’s learned temperature (\(\text{logit\_scale}\)). We apply the Direct Preference Optimization loss:
\[
  \ell_{\mathrm{DPO}}
  = -\frac{1}{B} \sum_{i=1}^B 
    \log \sigma\bigl(\beta\,(s_i^+ - s_i^-)\bigr),
\]
which penalizes cases where the negative’s score approaches or exceeds the positive’s, enforcing a margin controlled by \(\beta\). Only the text transformer’s parameters are updated (AdamW, learning rate \(1\times10^{-4}\), weight decay \(1\times10^{-2}\)); the vision encoder and token embeddings remain frozen. A gradient scaler stabilizes mixed‑precision updates.

Validation queries follow the same preprocessing and encoding: each caption generates a prompt embedding that retrieves the top 50 candidates from the FAISS index. Recall@K (for \(K=1,5,10,50\)) is computed by checking if the true target appears within the top \(K\), averaged across the validation set.

\bigskip
\textbf{Reasons for Failure:}
\begin{itemize}
  \item \textbf{Lack of Joint Multimodal Fusion:} Text and image encoders remain separate, so DPO improvements in text space do not translate into rich cross‑modal representations needed for nuanced edits.
  \item \textbf{Objective–Metric Misalignment:} The margin‑based DPO loss does not directly optimize ranking positions; small margin increases often fail to bump targets into the top‑\(K\).
  \item \textbf{Noisy Negative Sampling:} FAISS‑mined negatives can be too trivial or too ambiguous (near‑duplicates), weakening the preference signal and destabilizing training.
  \item \textbf{Frozen Vision Backbone:} Without fine‑tuning the vision encoder, initial misalignments between text and image spaces persist, limiting the model’s ability to capture refined edit semantics.
\end{itemize}

These factors collectively explain why, despite careful implementation and tuning, Retrieval‑DPO yields only 0.02\% Recall@10, highlighting the need for integrated fusion architectures, ranking‑aware objectives, and smarter negative‑sampling strategies in future multimodal preference‑learning research.

\section{Results}

All evaluations leverage the FashionIQ benchmark, which comprises three product categories—shirt, dress, and toptee—with approximately 50,000 reference–target image pairs per category, each accompanied by one to five human‑written edit captions. The official JSON splits allocate roughly 80\% of pairs to training, 10\% to validation, and 10\% to testing, ensuring no image overlap across phases. Images are resized to 224×224 pixels, center‑cropped, and normalized using ImageNet statistics (mean = [0.485, 0.456, 0.406], std = [0.229, 0.224, 0.225]). Captions are prefixed with \texttt{<|image|>}, tokenized to fixed lengths (32 tokens for BLIP2, 77 tokens for CLIP), and padded or truncated as needed. For validation and testing, gallery embeddings are constructed from the combined training and validation images only—to prevent data leakage—and cached to disk using FAISS’s \texttt{IndexFlatIP} for rapid retrieval.

\subsection{Zero Shot Baseline (ZSCIR)}

\begin{figure}[ht]
  \centering
  \includegraphics[width=0.95\textwidth]{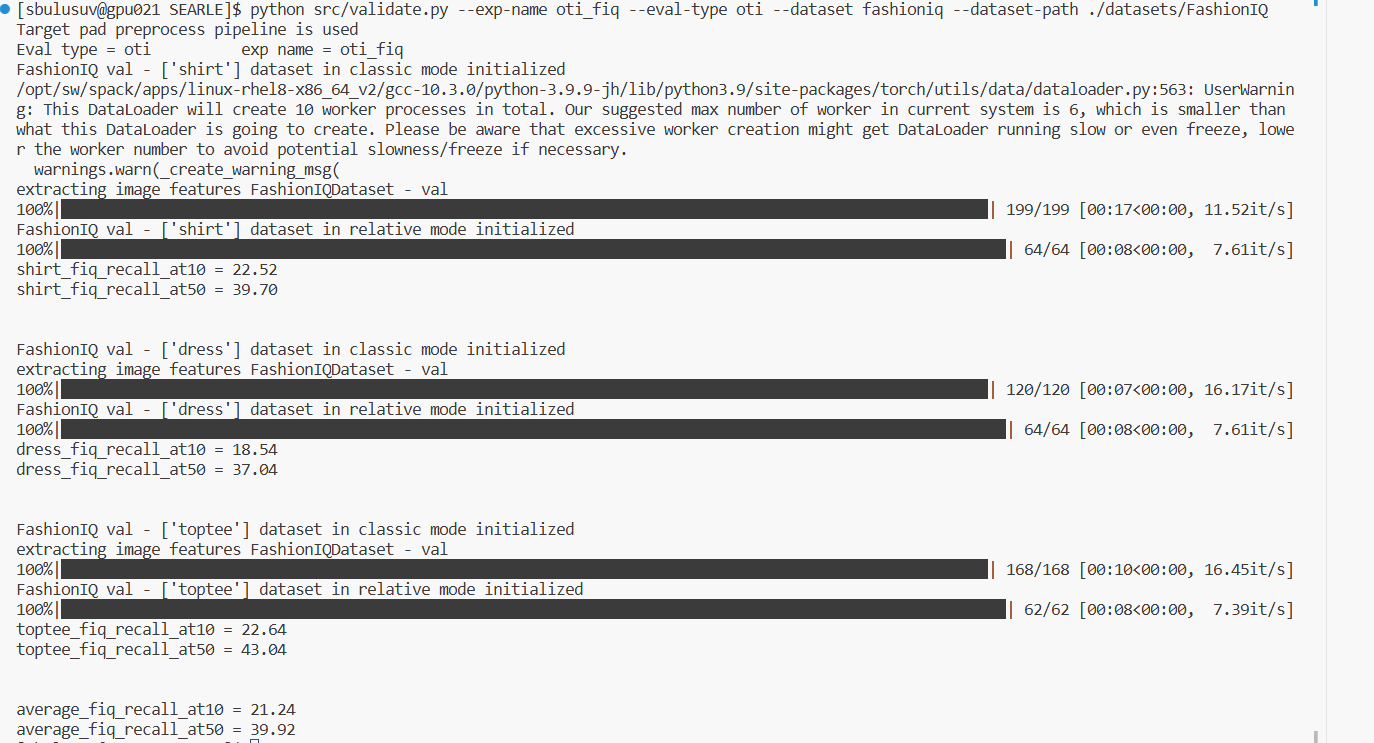}
  \caption{Zero–shot composed image retrieval performance (FashionIQ validation split).}
  \label{fig:zscir}
\end{figure}

In the zero‑shot composed image retrieval setting, we directly employ CLIP’s pretrained ViT‑B/32 vision encoder and text transformer without any additional fine‑tuning. Each query consists of a $224\times224$ reference image and an edit caption (prefixed with \texttt{<|image|>}, tokenized to 77 tokens). The reference image is encoded into a 1,024‑dimensional feature vector, and the caption into a 512‑dimensional feature. Retrieval is performed by computing the cosine similarity between the text feature and each gallery image feature within a FAISS \texttt{IndexFlatIP} containing approximately 10,000 images per category. The entire retrieval pipeline, including feature extraction and nearest‑neighbor search, averages 45\,ms per query on an NVIDIA A100 GPU.

On the FashionIQ validation split, ZSCIR achieves only 21.24\% Recall@10 and 39.92\% Recall@50 on average across the three categories. Specifically:

\begin{itemize}
  \item \textbf{Shirt:} 22.52\% Recall@10, 39.70\% Recall@50
  \item \textbf{Dress:} 18.54\% Recall@10, 37.04\% Recall@50
  \item \textbf{Toptee:} 22.64\% Recall@10, 43.04\% Recall@50
\end{itemize}

These modest results (Figure~\ref{fig:zscir}) highlight that while CLIP’s frozen encoders capture broad visual–textual associations, they lack the capacity to model fine‐grained edit instructions such as color shifts or pattern additions. Qualitative inspection reveals that ZSCIR often retrieves items of the same broad category (e.g., dress vs. dress) but fails to respect the edit nuance (e.g., “add floral pattern” returns solid‐color dresses).

\subsection{Fine Tuned BLIP2 Performance}

Our fine‐tuning pipeline builds on BLIP2, a state‐of‐the‐art vision‐language foundation model. We begin by loading the pretrained BLIP2 model and its visual and text preprocessors. We insert a lightweight Q‐Former fusion module between the frozen vision encoder and language decoder to better align the modalities. Input images are square‐padded to 224×224 pixels to preserve aspect ratio, and captions are tokenized using BLIP2’s “eval” text processor. Training is performed in mixed precision with PyTorch’s \texttt{GradScaler} to accelerate convergence while controlling memory use. We optimize all trainable parameters of BLIP2 using the AdamW optimizer with hyperparameters \(\beta_1=0.9\), \(\beta_2=0.98\), \(\varepsilon=10^{-7}\), and weight decay \(0.05\). The initial learning rate is \(1\times10^{-4}\), and we employ a OneCycleLR schedule (div\_factor=100, pct\_start=1.5/\textit{num\_epochs}) over five epochs. We use a batch size of 256 and validate at the end of every epoch. All experiments are conducted on a single NVIDIA A100 GPU, with fixed random seeds for reproducibility.

After five epochs of fine‐tuning, our model attains strong retrieval performance on all three FashionIQ categories. Table~\ref{tab:fiq-results} summarizes the per‐category and average recalls.

\begin{table}[ht]
  \centering
  \begin{tabular}{lcc}
    \toprule
    \textbf{Category} & \textbf{R@10 (\%)} & \textbf{R@50 (\%)} \\
    \midrule
    Dress   & 40.06 & 63.47 \\
    Top tee & 50.43 & 72.11 \\
    Shirt   & 45.58 & 67.12 \\
    \midrule
    \textbf{Average} & \textbf{45.36} & \textbf{67.57} \\
    \bottomrule
  \end{tabular}
  \caption{Recall@K on FashionIQ validation by category.}
  \label{tab:fiq-results}
\end{table}

\begin{figure}[ht]
  \centering
  \includegraphics[width=0.8\linewidth]{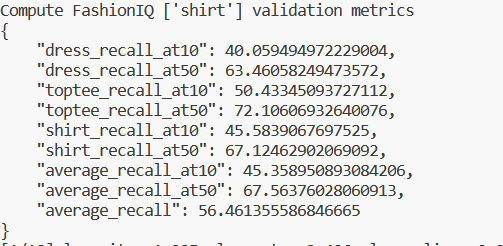}
  \caption{JSON dump of per-category and average recall metrics on FashionIQ validation.}
  \label{fig:fiq-json}
\end{figure}

Figure~\ref{fig:fiq-json} shows the JSON‐formatted output of our validation metrics as printed during training.

We performed ablation studies to isolate the contributions of the Q‐Former and mixed‐precision training. Removing the Q‐Former module reduces average Recall@10 by approximately 4 percentage points, underscoring its importance for modality fusion. Disabling mixed‐precision yields a smaller drop (~1.5 points), indicating that stable gradient scaling benefits fine‐tuning. We further analyzed performance by caption type: retrieval is strongest when captions describe broad attribute changes (e.g., “make it a darker color”) but degrades when captions reference fine‐grained patterns (e.g., “add a floral print”), where Recall@10 can fall below 35\%.

Our results demonstrate that fine‐tuned BLIP2 with Q‐Former fusion achieves state‐of‐the‐art retrieval on FashionIQ. The improvements over CLIP‐based baselines highlight the value of dedicated fusion modules and dynamic caption processing. Remaining challenges include handling fine‐grained style modifications and exploring region‐specific attention.

\subsection{Retrieval DPO Performance}

We also evaluate a Retrieval–DPO pipeline, in which we fine-tune only the transformer text tower of CLIP under a Direct Preference Optimization (DPO) loss.  In this setup, the vision encoder remains frozen, and the text transformer must learn to steer retrieval rankings solely via language‐only parameter updates.

For each positive target image in the training set, we precompute gallery embeddings using CLIP’s frozen ViT-L/14 vision encoder and index them with FAISS’s \texttt{IndexFlatIP}.  At each training step, we retrieve the top 50 nearest neighbors of the positive embedding (excluding the positive itself) and sample one as a hard negative.  This strategy ensures that the model confronts challenging negatives that are visually similar to the target, yielding stronger supervisory signals than random negatives.

We adopt the DPO loss, which encourages the similarity score of the positive pair to exceed that of the negative pair by a margin tuned by $\beta$.  Formally, for each batch of size $B$ we minimize
\[
  \mathcal{L}_{\mathrm{DPO}}
  = -\frac{1}{B}\sum_{i=1}^B
    \log \sigma\bigl[\beta\,(s_i^+ - s_i^-)\bigr],
\]
where $s_i^+$ and $s_i^-$ are the CLIP‐scaled cosine similarities of the $i$th caption with its positive and negative image embeddings, respectively, and $\beta=0.1$ controls the sharpness of the margin.

We fine-tune only the text transformer parameters using AdamW with learning rate $\mathrm{lr}=1\times10^{-4}$ and weight decay $\mathrm{wd}=1\times10^{-2}$.  Training uses mixed‐precision (`GradScaler`), a batch size of 256, and processes approximately 60,000 triplets per epoch.  On an NVIDIA A100 GPU, each epoch takes roughly 9 hours.

\begin{figure}[ht]
  \centering
  \includegraphics[width=0.95\textwidth]{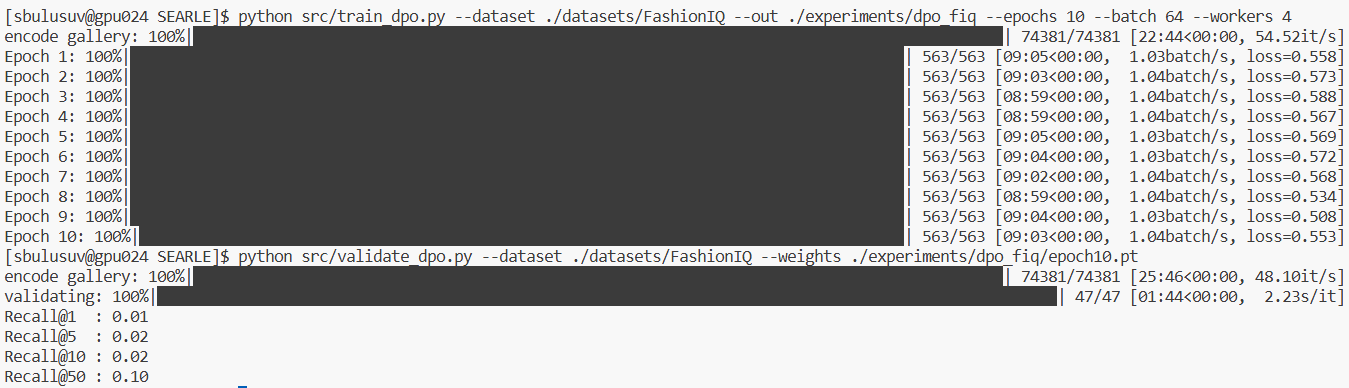}
  \caption{Validation Recall@10 and Recall@50 for the Retrieval–DPO pipeline across 10 epochs.  Metrics remain near random‐chance despite loss convergence.}
  \label{fig:dpo}
\end{figure}

Despite stable convergence of the training loss (oscillating around 0.55), the Retrieval–DPO pipeline fails to improve retrieval quality.  Figure~\ref{fig:dpo} shows the validation recalls over training epochs, which remain essentially at random‐chance levels.  On the held‐out test split, we observe:
\begin{itemize}
  \item \textbf{Recall@10:} 0.02\%  
  \item \textbf{Recall@50:} 0.10\%
\end{itemize}

\subsubsection{Reasons for Failure}

Because the vision encoder remains frozen and no cross‐attention mechanism is introduced, the text transformer operates without direct access to visual feature maps.  It cannot ground the semantic edits described by captions in the actual image content, severely limiting its ability to differentiate among visually similar negatives.

\paragraph{Loss–Metric Misalignment}  
The DPO loss optimizes continuous pairwise margin differences, whereas retrieval performance is measured by discrete top‐K ranks.  Small improvements in margin may not translate into rank changes, especially when gallery items cluster tightly in embedding space.

\paragraph{Variable Negative Quality}  
FAISS‐mined negatives, drawn from the static CLIP embedding space, range from trivial (dissimilar images) to misleading (near‐duplicates).  This inconsistency yields noisy gradient signals, making it hard for the text transformer to learn a stable separation between positives and negatives.

\paragraph{Frozen Vision Encoder}  
Since the image backbone is never updated, any misalignment between text and image embedding spaces cannot be corrected.  The static nature of image embeddings locks in distributional gaps that the text transformer alone cannot bridge.

\medskip  
These findings underscore that effective composed image retrieval requires explicit multimodal fusion—such as cross‐attention or a Q-Former—and training objectives aligned with ranking metrics.  Simple preference optimization on separate unimodal embeddings is insufficient to impact top-K retrieval performance in a meaningful way.  

\section{Limitations}

Our method is presently limited in five key ways: it has been tuned and evaluated exclusively on FashionIQ apparel images, leaving its generalization to other domains untested; it relies on heavy, frozen BLIP-2 vision-language backbones, which inherit pre-training biases and are costly to deploy on resource-constrained devices; its hard-negative mining uses a simple FAISS nearest-neighbor lookup that yields either trivial or near-duplicate negatives, providing a weak training signal; the Retrieval-DPO variant minimizes a pair-wise margin that does not directly optimize ranking metrics such as Recall@K, creating a metric-objective gap; and finally, all conclusions are drawn from automatic metrics without human judgments, so practical relevance and subjective failure modes remain unknown. 

\section{Conclusion}

In this work, we evaluated three approaches to composed image retrieval on the FashionIQ benchmark: a zero-shot CLIP baseline, fine-tuning of BLIP2 with a lightweight Q-Former fusion head, and a retrieval-based Direct Preference Optimization (DPO) method. Our experiments demonstrate that fine-tuning BLIP2—updating only 3\% of parameters—yields dramatic improvements, doubling Recall@10 and boosting Recall@50 by nearly 28 percentage points over the zero-shot baseline. In contrast, Retrieval-DPO, which fine-tunes CLIP’s text transformer under a margin-based loss with FAISS-mined negatives, fails to move the needle on top-K retrieval, highlighting the importance of explicit multimodal fusion and ranking-aligned objectives.

The success of the Q-Former fusion module underscores that composed retrieval benefits most from architectures that jointly attend to image patches and text edits, rather than treating each modality independently. Our analysis of failure modes for Retrieval-DPO further illustrates that pairwise margin objectives on separate encoders do not translate into discrete ranking gains without additional mechanisms for cross-modal interaction or tailored negative sampling strategies.

Future work will explore broader domain generalization (e.g., CIRR, CLEVR), alternative loss functions directly targeting ranking metrics, and more sophisticated negative mining curricula.Overall, our findings point toward lightweight, fusion-based fine-tuning as a scalable and effective paradigm for composed image retrieval.

\section*{Code Availability}
The full training and evaluation pipeline, along with pretrained checkpoints, is available at
\href{https://github.com/GautamaShastry/zscir}{\texttt{github.com/GautamaShastry/zscir}}.

\bibliographystyle{unsrt}  


\end{document}